\documentclass[conference]{IEEEtran}
\IEEEoverridecommandlockouts
\usepackage{amsmath,amsfonts}
\usepackage{algorithm}
\usepackage[noend]{algpseudocode}
\usepackage{hyperref}
\usepackage{multirow}
\usepackage{array}
\usepackage{textcomp}
\usepackage{caption}
\usepackage{subcaption}
\usepackage{stfloats}
\usepackage{url}
\usepackage{verbatim}
\usepackage{graphicx}
\def\BibTeX{{\rm B\kern-.05em{\sc i\kern-.025em b}\kern-.08em
    T\kern-.1667em\lower.7ex\hbox{E}\kern-.125emX}}
\usepackage{balance}
\begin{document}

\title{PMFL: Partial Meta-Federated Learning for heterogeneous tasks and its applications \\on real-world medical records}

\makeatletter 
\newcommand{\linebreakand}{%
  \end{@IEEEauthorhalign}
  \hfill\mbox{}\par
  \mbox{}\hfill\begin{@IEEEauthorhalign}
}
\makeatother 

\author{
\IEEEauthorblockN{Tianyi Zhang}
\IEEEauthorblockA{Fulton Schools of Engineering \\Arizona State University \\AZ, USA
\\Email: tzhan177@asu.edu}
\and
\IEEEauthorblockN{Shirui Zhang}
\IEEEauthorblockA{Singapore
\\Email: shirui\_zhang@outlook.com}
\and
\IEEEauthorblockN{Ziwei Chen}
\IEEEauthorblockA{Beijing Jiaotong University \\Beijing, China 
\\Email: ziwei8707@163.com}

\and
\linebreakand
\IEEEauthorblockN{Yoshua Bengio}
\IEEEauthorblockA{Mila - Quebec AI Institute \\CIFAR AI Chair \\QC, Canada}
\and
\IEEEauthorblockN{Dianbo Liu}
\IEEEauthorblockA{Broad Institute of MIT and Harvard \\MA, USA\\Mila - Quebec AI Institute \\QC, Canada
\\Email: liudianbo@gmail.com}

}

\maketitle
\begin{abstract}
 Federated machine learning is a versatile and flexible tool to utilize distributed data from different sources, especially when communication technology develops rapidly and an unprecedented amount of data could be collected on mobile devices nowadays. Federated learning method exploits not only the data but the computational power of all devices in the network to achieve more efficient model training. Nevertheless, while most traditional federated learning methods work well for homogeneous data and tasks, adapting the method to heterogeneous data and task distribution is challenging. This limitation has constrained the applications of federated learning in real-world contexts, especially in healthcare settings. Inspired by the fundamental idea of meta-learning, in this study we propose a new algorithm, which is an integration of federated learning and meta-learning, to tackle this issue. In addition, owing to the advantage of transfer learning for model generalization, we further improve our algorithm by introducing partial parameter sharing to balance global and local learning. We name this method partial meta-federated learning (PMFL). Finally, we apply the algorithms to two medical datasets. We show that our algorithm could obtain the fastest training speed and achieve the best performance when dealing with heterogeneous medical datasets. The source code is available at \href{https://github.com/destiny301/PMFL}{https://github.com/destiny301/PMFL}.
\end{abstract}


\section{Introduction} \label{introduction}
\IEEEPARstart{I}{n} our modern life, smartphones, tablets, and laptops are the primary computing devices, and phones are especially indispensable for many people \cite{McMahan2017CommunicationEfficient}. With the development of the processors and sensors of these devices, like cameras and microphones, an unprecedented amount of data would be produced when so many people carry them anywhere. Therefore, if we are able to exploit such data effectively and efficiently to train the neural networks, the performance of networks would be significantly improved. For instance, with the photos from the smartphones' cameras, we could train a better computer vision model to classify different images or segment different objects. With the text data, people enter into their devices, we could obtain a more robust language model to translate different languages. However, directly exploiting these data together is not feasible, because the data from mobile devices are usually private and significantly large in quantity, which means that it's insecure and impossible to gather all such data together to store in a centralized location (i.e. server). 

To address these issues, federated learning was proposed by researchers \cite{JK2016federated}.
In the frame of federated learning \cite{JK2016federated}, clients (e.g. different devices or locations) transmit the gradients of all model parameters, which are obtained from training local data of the clients, to the server instead of the original data. The server then aggregates the gradients with weighted average and updates the server model. After that, the server delivers the updated model to all clients for the next iterations. In another word, federated learning does not share data but only shares model parameters \cite{Bonawitz2017}. Thus, it reduces the privacy and security risks by containing the raw data within clients’ devices \cite{JK2016federated}. 
Despite that federated learning is capable of utilizing data from multiple clients to train the model, it only works well for homogeneous datasets but performs significantly worse for heterogeneous datasets whose training tasks are different. Nevertheless, datasets from different clients are usually heterogeneous in the real world. Specifically, the medical records of patients from different hospitals may be utilized to diagnose different diseases.
In comparison, while meta-learning, which exposes different training tasks to models for them to more productively learn new tasks, could handle the training of heterogeneous datasets, it's devised for local data training. 


Inspired by the basic idea of meta-learning, we propose a novel federated learning algorithm, which will feedback the training loss to the server rather than the gradients of model parameters, to deal with the heterogeneity of datasets from different clients and adapt federated learning to more general applications.
Furthermore, in the heterogeneous training scenario, the training task of the server (i.e. the global model) would be also different from all clients.
In order to make up for the heterogeneity between the server and clients and further enhance the performance and robustness of our algorithm, we freeze half of the global model and update only the other half with local data during the federated learning process to balance the global training and local training. The proposed algorithm experiments on two medical datasets, MIMIC-CXR Database v2.0.0 and eICU, and performs the best over various evaluation metrics compared to other federated learning algorithms.

\section{Related Work}
\textbf{Federated learning.}
While federated learning comes with unique statistical and systematical challenges that bottleneck the real-life application, like constraints in terms of storage, computation, and communication capacities \cite{JK2017federated, chen2019federated, 10.5555/3294996.3295196, zhao2018federated}, McMahan et al. \cite{McMahan2017CommunicationEfficient} proposed the Federated Averaging (FedAvg) algorithm that is robust to unbalanced and non-IID data distributions to achieve high accuracy. It also reduces the rounds of communication needed to train a deep network on decentralized data and strikes a good balance between computation and communication cost. 
They update the local model at first, and transmit the model parameters to the server, then applied the simple weighted average to obtain the global model parameters.
In addition, although model parameters learned in a traditional way still give out information about data used in training, more algorithms have been proposed to further protect the data privacy \cite{geyer2018differentially, Bonawitz2017, 10.1145/3298981, lin2020deep}.

\textbf{Meta-learning.}
Meta-learning is a powerful paradigm that enables machine learning models to learn how to learn over multiple learning episodes \cite{schmidhuber:1987:srl, Bengio97onthe, finn2018metalearning}. The learning algorithm usually covers a set of related tasks and improves its future learning performances with this experience gained \cite{finn2017modelagnostic, Rajeswaran2019MetaLearningWI, hospedales2020metalearning, Vanschoren2018}. 
Specifically, Vinyals et al. \cite{10.5555/3157382.3157504} used Matching Network as meta-learner that applied non-parametric structure to remember and adapt to new training sets. Santoro et al. \cite{10.5555/3045390.3045585} came up with Memory-Augmented Neural Networks (MANNs) that utilize Neural Turing Machines (NTMs) \cite{graves2014neural} as a meta-learner to rapidly learn from new data and make accurate predictions with only a few samples. Ravi and Larochelle \cite{Ravi2017OptimizationAA} proposed an LSTM-based meta-learner that uses its state to represent the learning updates of a classifier’s parameters. Importantly, Finn et al. \cite{finn2017modelagnostic} proposed a model-agnostic meta-learning (MAML) algorithm that is applicable to any model trained with gradient descent, which includes classification, regression, and reinforcement learning \cite{finn2017modelagnostic}. The model’s initial parameters are trained such that good generalization performance can be achieved with a small number of gradient steps with a limited amount of training data \cite{finn2017modelagnostic, hospedales2020metalearning}. Models learned with MAML use fewer parameters as compared to matching networks and the meta-learner LSTM and outperform memory-augmented networks \cite{10.5555/3045390.3045585, pmlr-v70-munkhdalai17a} and the meta-learner LSTM. Therefore, MAML is definitely the best option for our purpose.

\textbf{Applications in medicine.}
Because privacy protection is crucial in healthcare and medical research, various research tasks have explored applications of federated learning in healthcare, e.g. patient similarity learning, patient representation learning, phenotyping, and predictive modeling \cite{Sadilek2020.12.22.20245407, info:doi/10.2196/20891, boughorbel2019federated, xu2020federated, liu2022confederated}. Typical research works deal with EHR/ICU/Genomics data as well as image and sensor data \cite{10.1145/3412357}. Under a federated setting, Lee et al. \cite{osti_10064144} proposed a privacy-preserving platform for patient similarity learning across institutions. They used k-NN model based on hashed EHRs and sought to identify similar patients from one hospital to another. Huang and Liu \cite{huang2019patient} proposed community-based federated machine learning (CBFL) algorithm to cluster the distributed data into clinically meaningful communities and evaluated it on non-IID ICU data. Liu et al. \cite{liu2019twostage} developed a two-stage federated natural language processing method that utilizes clinical notes from different hospitals and clinics and conducted patient representation learning and obesity comorbidity phenotyping tasks in a federated manner. The result shows the federated training of ML models on distributed data outperforms algorithms training on data from a single site. Liu et al. \cite{liu2018fadlfederatedautonomous} suggested a Federated-Autonomous Deep Learning (FADL) approach that predicts mortality of patients without moving health data out of their silos using distributed machine learning approach. Brisimi et al. \cite{BRISIMI201859} predicted future hospitalizations for patients using EHR data spread among various data sources/agents by training a sparse Support Vector Machine (sSVM) classifier in federated learning environment.

Additionally, for medical image classification, most of the state-of-the-art deep learning methods require a sufficient amount of training data \cite{jimaging7020031, park2021metalearning}. However, in the medicine domains, it is very expensive to annotate the object of interest, especially true for rare or novel diseases \cite{Hu2018MetaLearningFM, 9150592}. Under this context, meta-learning is very practical in addressing such limitations \cite{nichol2018firstorder, park2021metalearning, Hu2018MetaLearningFM}. Hu et al. \cite{Hu2018MetaLearningFM} implemented Reptile, a state-of-the-art meta-learning model pre-trained with mini-ImageNet, for detecting diabetic retinopathy. It simplifies MAML and makes it more scalable. Park et al. \cite{park2021metalearning} applied meta-learning in medical image registration. Zhang et al. \cite{jimaging7020031} proposed an optimization-based meta-learning method for medical image segmentation tasks. The algorithm is able to surpass MAML and Reptile in terms of the Dice similarity coefficient (DSC) with U-Net being the baseline model. Khadga et al. \cite{khadga2021fewshot} used Implicit Model Agnostic Meta Learning (iMAML) algorithm in few-shot learning setting for medical image segmentation. Chen et al. \cite{chen2021metadelta} proposed MetaDelta, a novel and practical meta-learning system for the few-shot image classification.

\section{Methodology}
As mentioned in section \ref{introduction}, while traditional federated learning could handle many clients with the same training task, it couldn't be applied to heterogeneous training tasks which happen more frequently in the real world. To address this issue, we propose a novel federated learning algorithm based on meta-learning, which feedback the training loss rather than gradients of model parameters from clients to the server, as shown in Fig. \ref{fig:intro}. In this section, the detailed algorithm, Meta-Federated Learning (MetaFL), which effectively integrates federated learning and meta-learning, will be introduced in \ref{metafl}. Moreover, in \ref{pmfl}, we will talk in detail about our optimal solution, Partial Meta-Federated Learning (PMFL), which further enhances MetaFL on the basis of transfer learning.

\begin{figure}[H]
\centering
\begin{subfigure}[b]{\columnwidth}
    \centering
    \includegraphics[width=\columnwidth]{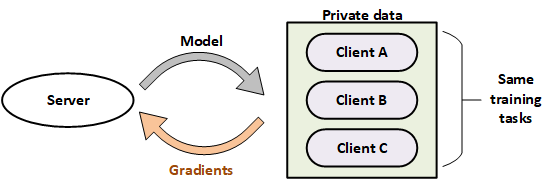}
    \caption[width=\columnwidth]{Conventional federated learning process.}
    \label{fig:intro1}
\end{subfigure}
\hfill
\begin{subfigure}[b]{\columnwidth}
    \centering
    \includegraphics[width=\columnwidth]{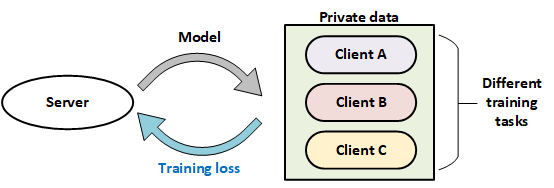}
    \caption[width=\columnwidth]{Proposed method for heterogeneous datasets.}
    \label{fig:intro2}
\end{subfigure}
\caption{Comparison of the conventional federated learning and our proposed algorithm for heterogeneous datasets.}
\label{fig:intro}
\end{figure} 

\subsection{Meta-Federated Learning (MetaFL)} \label{metafl}
With the benefits of federated learning which could leverage the computation resources from all clients and avoid the privacy issues of users' data, we are able to train a model from a set of clients that provide different data. However, with federated learning, we just simply adopt more data for a common task and the method would not work effectively for heterogeneous datasets from different clients. Here, different from the non-IID setting with identical tasks but different distributions, the heterogeneity means there are different tasks in different silos (i.e. clients). On the other hand, meta-learning, such as Model-Agnostic Meta-Learning (MAML) \cite{finn2017modelagnostic}, aims at training an adaptive model and has the ability to deal with different tasks, but it is proposed for local data learning and lacks the capacity of training from remote data. 

While federated learning could safely train the model with datasets from different remote clients, MAML has the capacity of training an adaptive model with heterogeneous datasets. Therefore, if we could borrow some insights from MAML to design an adaptive federated learning algorithm, we would successfully train our model with heterogeneous remote datasets\cite{fallah2020personalized}.

Actually, the fundamental structures of federated learning and MAML are grossly similar. Specifically, for federated learning, such as FederatedAveraging (FedAvg) algorithm \cite{McMahan2017CommunicationEfficient}, which reduces the server's computation cost by updating models on local clients and feedbacking updated model parameters to the server, each client needs to update the model parameters with the local dataset after receiving the model from the server, then send all the model parameters back to the server and the server will integrate them with weighted average based on the number of samples assigned to different clients as the updated model which would be delivered to all clients in next iterations. The following equations show the major work,
\begin{equation} \label{eq:1}
    \theta_k \leftarrow \theta - \eta g_k,\; \forall k
\end{equation}
\begin{equation}
    \theta \leftarrow \sum_{k = 1}^K \frac{n_k}{n}\theta_k,
\end{equation}
where $\eta$ is the learning rate, $K$ is the number of clients, $g_k$ is the gradient for client $k$, $\theta_k$ indicate the parameters for the $kth$ client, $\theta$ indicate the parameters of the server, $n_k$ is the number of samples for the $kth$ client, and $n$ is the total number of samples for all clients. Notice, for equation \eqref{eq:1}, each client could iterate the local update for multiple epochs before transmitting the model parameters to the server.

Whilst for MAML algorithm \cite{finn2017modelagnostic}, the model could be represented by a parametrized function $f_\theta$ with parameters vector $\theta$, and for each training task $k$, we need to compute the losses $\mathcal{L}_k(f_\theta)$ using samples assigned to this training task and integrate all the losses. Importantly, our main model keeps unchangeable during this meta training process, which is similar to FedAvg in which the global model would not change during the clients' updates. After that, we utilize the integrated loss to backward and update our main model parameters $\theta$ as the following equations,
\begin{equation} \label{eq:3}
    \theta_k \leftarrow \theta - \alpha \nabla_\theta \mathcal{L}_k(f_\theta),\; \forall k
\end{equation}
\begin{equation} \label{eq:4}
    \theta \leftarrow \theta - \beta \nabla_\theta \sum_{k=1}^K \frac{1}{K}\mathcal{L}_k(f_{\theta_k}),
\end{equation}
where $\alpha$ is the learning rate for meta training models, $\beta$ is the learning rate for our main model, $\theta_k$ indicates the parameters for the $kth$ training task, $\theta$ indicate the parameters for the main model. Similarly, for equation \eqref{eq:3}, each training task could iterate the meta gradient descent step for multiple epochs before the loss integration. And, for equation \eqref{eq:4}, we can know that the main model parameters $\theta$ would be trained by optimizing the integrated loss which is based on various meta training model parameters $\theta_k$.

For both of FedAvg and MAML, we need to train various submodels with different data at first and integrate the training results of submodels to update the main model. To be specific, we need to train the clients' models with the local datasets and directly integrate the model parameters as the new global model for FedAvg \cite{McMahan2017CommunicationEfficient}, while we need to finish the meta-training of different tasks and integrate the training losses to update the main model for MAML \cite{finn2017modelagnostic}. 

Inspired by the mechanism of MAML which could handle heterogeneous training tasks by leveraging the meta training loss, we propose an adaptive federated learning algorithm, Meta-Federated Learning (MetaFL), which differs from the traditional federated learning algorithms mainly by returning the losses $\mathcal{L}_k(f_\theta)$ from every client $k$ to the server, instead of model parameters $\theta_k$ or model gradients $g_k$, to considerably improve the performance of federated learning for heterogeneous remote data.
In MetaFL, the clients first utilize their own local data to update their models received from the server as equation \eqref{eq:3} for a few epochs then return the losses of the last epoch in the local training to the server.
After receiving all the losses, the server would compute the average of them and exploit the integrated loss to update the server model as equation \eqref{eq:4}. Subsequently, the server would start the new iteration by distributing the new model to all clients and waiting for them to return the new local losses for this iteration. 
Owing to the similar scheme to MAML in which the main model is updated with the training losses of submodels rather than explicit model parameters, MetaFL could be applied to more scenarios where clients may need to resolve different training tasks.
The specific MetaFL algorithm is depicted in Algorithm \ref{alg:metaFL}.


\begin{algorithm}
\caption{Meta-Federated Learning}\label{alg:metaFL}
\begin{algorithmic}[1]
\State randomly initialize $\theta$
\For{each round t = 1, 2, ...}
\State Sample K training tasks from various clients
\For{all K clients \textbf{in parallel}}
\State Evaluate $\nabla_\theta \mathcal{L}_k(f_\theta)$
\State Compute adapted local parameters with gradient descent: $\theta_k = \theta - \alpha \nabla_\theta \mathcal{L}_k(f_\theta)$ 
\EndFor
\State \textbf{end for}
\State clients return all losses $\mathcal{L}_k(f_{\theta_k})$ to the server
\State update the server model with the losses from all clients: $\theta \leftarrow \theta - \beta \nabla_\theta \sum_{k = 1}^K \frac{1}{K}\mathcal{L}_k(f_{\theta_k})$
\EndFor
\State \textbf{end for}
\end{algorithmic}
\end{algorithm}

\subsection{Partial Meta-Federated Learning (PMFL)} \label{pmfl}
\begin{figure}[ht]
\centering
\includegraphics[width=\columnwidth]{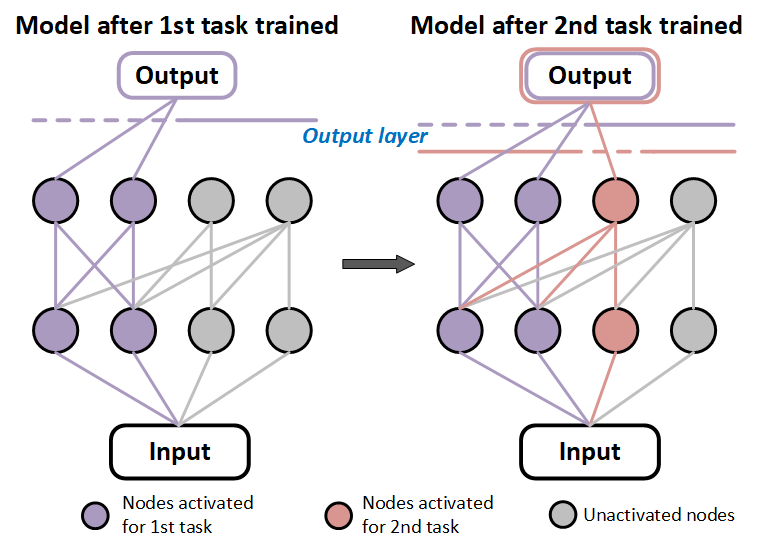}
\caption{\textbf{Continual Learning:} A simple example of continual learning shows its sparsification scheme.\label{fig:transfer}}
\end{figure}
Because the task that the server needs to settle is also divergent from the clients' and this heterogeneity would harm the final performance, transfer learning which focuses on multi-task learning and model generalization is an ideal utensil to further improve our algorithm. In particular, continual learning, a specific mechanism of transfer learning, could leverage a pre-trained model to train a new task beyond what has been previously trained \cite{yoon2017lifelong, golkar2019continual}. Owing to the characteristic of over-parametrization for neural networks, continual learning proposed a sparsification scheme that we just use some active nodes to train the specific task with graceful forgetting. In this case, our model would not suffer any catastrophic forgetting issues. Figure. \ref{fig:transfer} shows a simple example of continual learning. The subsequent tasks would use the unused weights (i.e. unactivated nodes) of the neural network to learn features since the sparsification scheme allows the current task to train models with only a subset of neurons. To be specific, the second task would be learned with the red neural nodes besides the purple ones while the purple nodes have been utilized for the training of the first task.

\begin{figure*}[ht]
\centering
\includegraphics[width=6.5in]{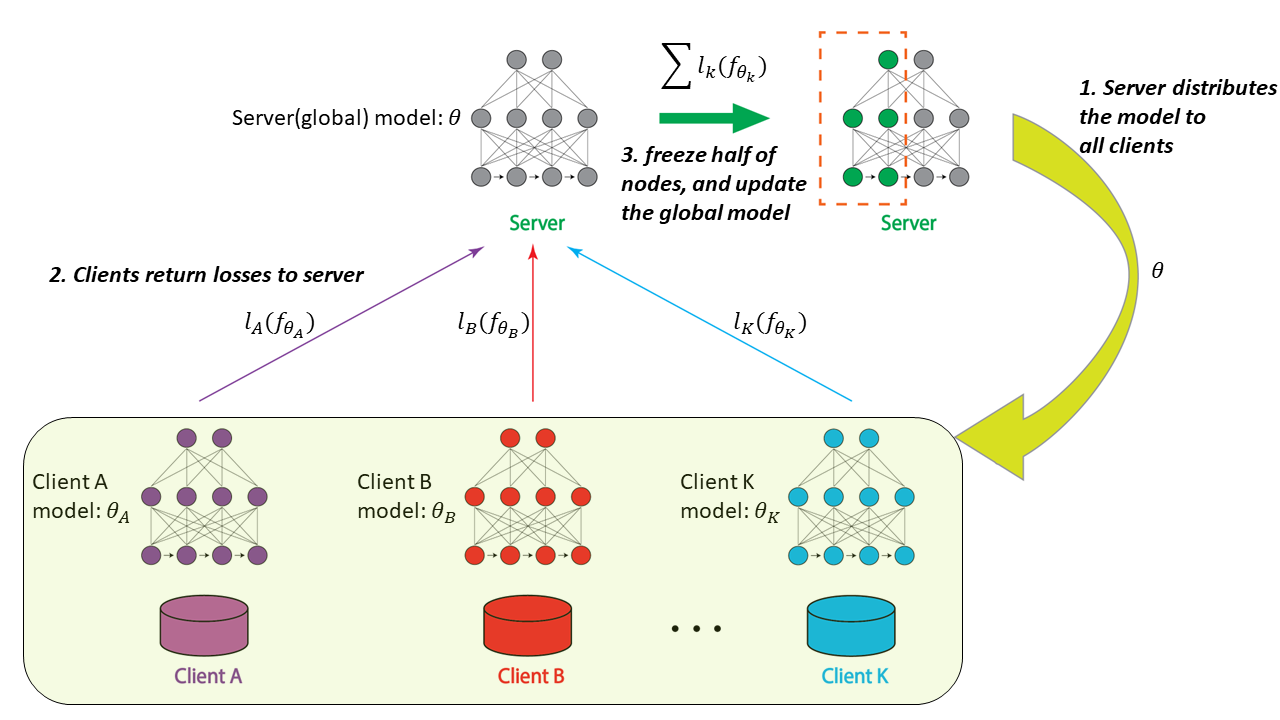}
\caption{\textbf{Flowchart of Partial Meta-Federated Learning (PMFL):} Similar to MetaFL, the server distributes the main model to all clients, then the clients would train with the local datasets for a few epochs. Then, all the losses from the last epoch of their local training would be returned to the server. After all the losses are delivered to the server, the server would freeze half of the main model and update the other half with the average loss which is computed with the returned losses from all clients.\label{fig:pmfl}}
\end{figure*}

Similarly, in our federated learning scenario, we could deactivate partial nodes of the neural network during training to avoid catastrophic forgetting.
In accordance with this thought, we further modify and optimize our method and propose a superior algorithm, Partial Meta-Federated Learning (PMFL). In PMFL, we take advantage of the over-parametrization of neural networks by using an activation-based neural pruning sparsification scheme to train models which only use a fraction of their width. Explicitly, we would activate the first half of neural nodes for all clients' training, which offers us optimal improvement in our experiments. These procedures of PMFL are visualized in Fig. \ref{fig:pmfl}. Let N denote the total number of nodes in our neural network. Specifically, denoting the first half of our main model parameters $\theta$ with $\theta[:\frac{N}{2}]$ and the second half with $\theta[\frac{N}{2}:]$, our PMFL algorithm would select $\theta[:\frac{N}{2}]$ to update when training with the integrated loss from all clients, and keep the other half of $\theta[\frac{N}{2}:]$ frozen and unchangeable. And all the clients would still update all model parameters $\theta_k$ after receiving the half model, then feedback all the losses of the last epoch to the server. By this, the server model has the ability to learn the common features from all various training tasks, and all the clients' model could keep their own features. Also, there would be some inactive nodes for our main model to learn future tasks.
The steps of the full PMFL algorithm are outlined in Algorithm \ref{alg:PMFL}.

\begin{algorithm}
\caption{Partial Meta-Federated Learning}\label{alg:PMFL}
\begin{algorithmic}[1]
\State randomly initialize $\theta, \theta_1, \theta_2, ..., \theta_K$ \Comment{$\theta$ indicates the server model, and $\theta_1, \theta_2, ..., \theta_K$ indicate K client models}
\For{each round t = 1, 2, ...}
\State Sample K training tasks from various clients
\For{all K clients \textbf{in parallel}} \Comment{each client could update locally in multiple epochs}
\State send the first half of the server model parameters to the client model: $\theta_k \leftarrow \theta[:\frac{N}{2}]$
\State Evaluate $\nabla_{\theta_k} \mathcal{L}_k(f_{\theta_k})$
\State Compute adapted local parameters with gradient descent: $\theta_k \leftarrow \theta_k - \alpha \nabla_{\theta_k} \mathcal{L}_k(f_{\theta_k})$ 
\EndFor
\State \textbf{end for}
\State clients return all losses $\mathcal{L}_k(f_{\theta_k})$ to the server
\State freeze the second half of the server model $\theta[\frac{N}{2}:]$
\State server update with gradient descent: $\theta \leftarrow \theta - \beta \nabla_\theta \sum_{k = 1}^K \frac{1}{K}\mathcal{L}_k(f_{\theta_k})$ \Comment{update the server model}
\EndFor
\State \textbf{end for}
\end{algorithmic}
\end{algorithm}

\section{Experimental Results}
To demonstrate the advantage of our algorithm, PMFL, we compare it with three baselines, training directly without any pretraining process (w/o FL) which denotes the plain model training and serves as the most fundamental baseline in our experiments, traditional federated learning (FL) is exactly the same with the FedAvg algorithm in \cite{McMahan2017CommunicationEfficient}, and MetaFL whose implementation is based on the algorithm in \cite{fallah2020personalized}.
Divergent from the setting of FL, for both of our MetaFL and PMFL algorithms, we arrange that the last client which possesses its own dataset represents the server for our scheme while the server for FL doesn't include any data and only works for integrating and updating the model. Similar to MAML, after we got the pre-trained model with data from all clients by MetaFL or PMFL algorithm, we would continue to train the model with part of the dataset from this server and test the final model with the other part. Thus, we would test our trained model with data from the server while training data consists of the datasets from all different clients. It's because of this arrangement for the server that we are able to train our model directly with the server's own data regardless of the data of clients to be the first baseline (w/o FL). Totally, there are four schemes to be compared in our experiments.

In order to completely assess model performance, we need to evaluate our algorithm with different metrics. We experiment on two medical datasets, and the model attempts to classify whether patients have certain diseases or not. After comparing our predictions with the true labels in which positive means confirmed case, all results would be split into four classes, which are true positive (TP), false positive (FP), true negative (TN), and false negative (FN). After we know how many there are for each class, we could easily derive some crucial metrics, such as recall (TPR), precision, and $ F_1$ score as the following equations:
\begin{equation} \label{eq:5}
    recall(TPR) = \frac{TP}{TP+FN}
\end{equation}
\begin{equation} \label{eq:6}
    precision = \frac{TP}{TP+FP}
\end{equation}
\begin{equation} \label{eq:8}
    F_1 = \frac{2}{recall^{-1}+precision^{-1}}
\end{equation}

The outcomes generated from our model are positive probabilities of the patients. We select a threshold to decide whether the result is positive or otherwise negative for each prediction.
By choosing different thresholds, we would get a different number of positive predictions, which would result in different precision, recall and so on. The ROC curve was produced by plotting the true positive rate (TPR or recall) and the false positive rate (FPR or 1-specificity) at thresholds ranging from 0 to 1. And, ROC AUC which we evaluate is the area under the ROC curve. Because the precision or other scores will be different if we select different thresholds, unlike ROC AUC, we couldn't directly exploit these scores to make numeric comparisons and conclude which algorithm is the best. Notwithstanding, the Youden index, the sum of sensitivity and specificity minus one, is an index used for setting optimal thresholds on medical tests, which could provide the best tradeoff between sensitivity and specificity. 

\begin{equation} \label{eq:9}
    J = TPR - FPR
\end{equation}
where we denote Youden index with $J$.

Therefore, by selecting the threshold to maximize the Youden index as equation \eqref{eq:9}, we would obtain the optimum cut-off point when our diagnostic test with the metrics, like Precision, Recall, and $F_{1}$ score,  gives a numeric rather than a dichotomous result.



\subsection{MIMIC-CXR v2.0.0}
\textbf{Dataset.}
The MIMIC Chest X-ray (MIMIC-CXR) Database v2.0.0 is a large publicly available dataset of chest radiographs in DICOM format with free-text radiology reports and structured labels derived from these reports. To be specific, the dataset contains 377,110 JPG format images and structured labels derived from the 227,827 free-text radiology reports associated with these images \cite{johnson2019mimic, johnson2019mimicpaper, goldberger2000physiobank, johnsonmimic}. In our experiment, we would mainly use two files from MIMIC-CXR v2.0.0: mimic-cxr-reports.tar.gz and mimic-cxr-2.0.0-chexpert.csv.gz.

To confirm the extraordinary performance of our proposed algorithm for heterogeneous training tasks, we split the entire dataset into multiple subsets which include different labels to represent different clients. For MIMIC-CXR v2.0.0 dataset, the basic model for all clients and the server would use the text reports of patients from \textit{mimic-cxr-reports.tar.gz} as input data to classify whether these patients have certain diseases, so the labels for different clients are corresponding to different columns from \textit{mimic-cxr-2.0.0-chexpert.csv.gz}, and each column of this file includes the information for a single kind of disease. Specifically, we take only 8 kinds of diseases, which are relevant to the lung, such as \textit{Pneumonia}, \textit{Lung Lesion}, \textit{Pneumothorax} and so on, into consideration and generate our 8 heterogeneous training tasks as shown in Fig. \ref{fig:mean}. It closely resembles the real-world scenario of federated learning and meta-learning. 


We extract 8 silo datasets without overlapping from the original dataset by making sure:
\begin{itemize}
    \item the 1:1 ratio of positive samples and negative samples for each silo.
    \item first extract data from the class which includes more samples to generate 8 different datasets with similar size in case some clients hugely dominate.
\end{itemize}

After that, we shuffle all silos to simulate 8 different clients of FedAvg (FL), MetaFL, or PMFL. During the experiment, we would choose one silo as the target task for evaluation, and randomly select any other 5 silos as training tasks to pretrain our model for FL, MetaFL and PMFL. For all clients, 90\% of the data would be used for training while 10\% would be used for testing.

\textbf{Model and performance.}
We could separately set different hyperparameters for the training of clients' models and the server model. 
To be specific, because our input data is text, our client model is based on LSTM, and it includes one embedding layer (embedding dimension of 128), one bidirectional LSTM (32 hidden nodes) and one linear layer with one output node. The binary loss function trained with Adam optimizer is adopted to evaluate the result. We settle on the meta learning rate of $10^{-3}$ and set batch size to be varied with respect to the size of local dataset in order to make sure that the number of batches for different clients would be identical, and the number of epochs is $10$. For our server model, the learning rate is $10^{-2}$.

\begin{table*}[hb]
\caption{Compare the AUC of ROC, Precision, Recall and $ F_1$ score for four different training cases when treating different diseases as target test tasks. Notice that Precision, Recall and $ F_1$ score are obtained with the thresholds which maximize Youden's Index.}
\centering
\begin{tabular}{l l l l l l}
\hline
Test task & Algorithm & ROC AUC & Precision & Recall & $ F_1$ score \\
\hline
\multirow{4}{6em}{Atelectasis}
& w/o FL & $0.9920\pm0.0027$ & $0.9709\pm0.0127$ & $0.9872\pm0.0036$ & $0.9789\pm0.0054$\\

& FL & $0.9857\pm0.0097$ & $0.9578\pm0.0219$ & $0.9756\pm0.0107$ & $0.9666\pm0.0159$\\

& MetaFL & $0.9923\pm0.0028$ & $0.9661\pm0.0124$ & $0.9845\pm0.0067$ & $0.9752\pm0.0083$\\

& \textbf{PMFL(Our method)} & $\mathbf{0.9971\pmb{\pm}0.0005}$ & $\mathbf{0.9863\pmb{\pm}0.0015}$ & $\mathbf{0.9899\pmb{\pm}0.0015}$ & $\mathbf{0.9881\pmb{\pm}0.0005}$\\
\hline
\multirow{4}{6em}{Consolidation}
& w/o FL & $0.9475\pm0.0347$ & $0.9268\pm0.0270$ & $0.9087\pm0.0832$ & $0.9160\pm0.0523$\\

& FL & $0.9475\pm0.0408$ & $0.9148\pm0.0631$ & $0.9284\pm0.0351$ & $0.9204\pm0.0415$\\

& MetaFL & $0.9690\pm0.0208$ & $0.9593\pm0.0242$ & $0.9148\pm0.0757$ & $0.9343\pm0.0398$\\

& \textbf{PMFL(Our method)} & $\mathbf{0.9900\pmb{\pm}0.0041}$ & $\mathbf{0.9635\pmb{\pm}0.0096}$ & $\mathbf{0.9737\pmb{\pm}0.0074}$ & $\mathbf{0.9686\pmb{\pm}0.0084}$\\
\hline
\multirow{4}{6em}{Lung Lesion}
& w/o FL & $0.8733\pm0.1054$ & $0.8621\pm0.1125$ & $0.8083\pm0.1193$ & $0.8339\pm0.1146$\\

& FL & $0.9287\pm0.0267$ & $0.9072\pm0.0327$ & $0.8853\pm0.0628$ & $0.8945\pm0.0352$\\

& MetaFL & $0.9713\pm0.0171$ & $0.9372\pm0.0249$ & $0.9570\pm0.0225$ & $0.9469\pm0.0230$\\

& \textbf{PMFL(Our method)} & $\mathbf{0.9817\pmb{\pm}0.0036}$ & $\mathbf{0.9505\pmb{\pm}0.0078}$ & $\mathbf{0.9653\pmb{\pm}0.0051}$ & $\mathbf{0.9578\pmb{\pm}0.0052}$\\
\hline
\multirow{4}{6em}{Lung Opacity}
& w/o FL & $0.9041\pm0.0152$ & $0.8426\pm0.0375$ & $0.9154\pm0.0711$ & $0.8745\pm0.0245$\\

& FL & $0.8830\pm0.0356$ & $0.8186\pm0.0511$ & $0.9067\pm0.0495$ & $0.8583\pm0.0293$\\

& MetaFL & $0.9669\pm0.0156$ & $0.9339\pm0.0239$ & $0.9575\pm0.0182$ & $0.9455\pm0.0187$\\

& \textbf{PMFL(Our method)} & $\mathbf{0.9882\pmb{\pm}0.0029}$ & $\mathbf{0.9605\pmb{\pm}0.0023}$ & $\mathbf{0.9723\pmb{\pm}0.0100}$ & $\mathbf{0.9663\pmb{\pm}0.0059}$\\
\hline
\multirow{4}{6em}{Pleural Effusion}
& w/o FL & $0.9436\pm0.0280$ & $0.8957\pm0.0310$ & $0.9208\pm0.0783$ & $0.9063\pm0.0127$\\
& FL & $0.9428\pm0.0111$ & $0.8832\pm0.0202$ & $0.9480\pm0.0271$ & $0.9142\pm0.0163$\\
& MetaFL & $0.9794\pm0.0057$ & $0.9436\pm0.0178$ & $0.9831\pm0.0076$ & $0.9629\pm0.0124$\\
& \textbf{PMFL(Our method)} & $\mathbf{0.9935\pmb{\pm}0.0015}$ & $\mathbf{0.9732\pmb{\pm}0.0086}$ & $\mathbf{0.9884\pmb{\pm}0.0051}$ & $\mathbf{0.9807\pmb{\pm}0.0044}$\\
\hline
\multirow{4}{6em}{Pleural Other}
& w/o FL & $0.8702\pm0.0544$ & $0.8334\pm0.0754$ & $0.8304\pm0.0777$ & $0.8297\pm0.0640$\\
& FL & $0.7166\pm0.0671$ & $0.6665\pm0.0541$ & $0.7471\pm0.1110$ & $0.7003\pm0.0694$\\
& MetaFL & $0.9489\pm0.0554$ & $0.9230\pm0.0941$ & $0.9451\pm0.0374$ & $0.9323\pm0.0643$\\
& \textbf{PMFL(Our method)} & $\mathbf{0.9799\pmb{\pm}0.0036}$ & $\mathbf{0.9656\pmb{\pm}0.0109}$ & $\mathbf{0.9578\pmb{\pm}0.0239}$ & $\mathbf{0.9615\pmb{\pm}0.0125}$\\
\hline
\multirow{4}{6em}{Pneumonia}
& w/o FL & $0.9394\pm0.0230$ & $0.8802\pm0.0255$ & $0.9305\pm0.0392$ & $0.9041\pm0.0240$\\
& FL & $0.9089\pm0.0349$ & $0.8392\pm0.0460$ & $0.9375\pm0.0229$ & $0.8849\pm0.0288$\\
& MetaFL & $0.9621\pm0.0191$ & $0.9123\pm0.0139$ & $0.9430\pm0.0396$ & $0.9270\pm0.0238$\\
& \textbf{PMFL(Our method)} & $\mathbf{0.9798\pmb{\pm}0.0084}$ & $\mathbf{0.9234\pmb{\pm}0.0082}$ & $\mathbf{0.9727\pmb{\pm}0.0067}$ & $\mathbf{0.9473\pmb{\pm}0.0039}$\\
\hline
\multirow{4}{6em}{Pneumothorax}
& w/o FL & $0.9725\pm0.0044$ & $0.9519\pm0.0079$ & $0.9543\pm0.0204$ & $0.9529\pm0.0089$\\
& FL & $0.9544\pm0.0129$ & $0.9030\pm0.0223$ & $0.9343\pm0.0277$ & $0.9180\pm0.0172$\\
& MetaFL & $0.9801\pm0.0049$ & $0.9613\pm0.0052$ & $0.9672\pm0.0155$ & $0.9642\pm0.0086$\\
& \textbf{PMFL(Our method)} & $\mathbf{0.9887\pmb{\pm}0.0020}$ & $\mathbf{0.9700\pmb{\pm}0.0028}$ & $\mathbf{0.9781\pmb{\pm}0.0077}$ & $\mathbf{0.9740\pmb{\pm}0.0035}$\\
\hline
\end{tabular}

\label{tab:2}
\end{table*}

In addition, to test the performance of our algorithm, we need to create a test model for the server, which would load the trained server model to start training, with one silo of data that we don't use for training before. And the hyperparameters are totally identical with the training clients, besides the batch size of $64$. All models were trained on a single NVIDIA 1080 TI GPU.

As we stated before, evaluation metrics include not only ROC AUC, but also the precision, recall, $ F_1$ score when maximizing the Youden index. By selecting different clients to be training tasks or test task, we compare our PMFL algorithm with three cases, which are training directly without any pretraining process (w/o FL), training with the pre-trained model from the FedAvg Learning algorithm (FL), and training with the pre-trained model from the MetaFL algorithm (MetaFL). 
\begin{figure}[ht]
\centering
\includegraphics[width=\columnwidth]{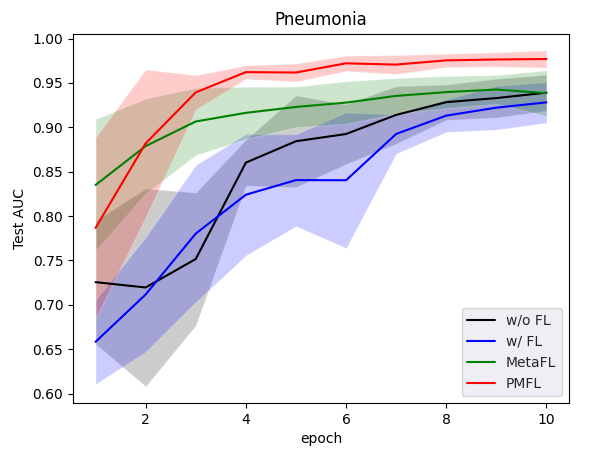}
\caption{Learning curve on average of our PMFL algorithm, against without FL, original FL, MetaFL when choosing Pneumonia as test task. The lines in this figure show the average AUC of 5 experiments while the thickness of the lines shows the standard deviation for each case.}
\label{fig:auc}
\end{figure}

Figure. \ref{fig:auc} shows the learning curve when Pneumonia was chosen as test task and evaluated by the AUC of ROC. Furthermore, in Fig. \ref{fig:mean}, we show the AUC mean of four algorithms for all these 8 server tasks. Three major conclusions are conveyed by the figures. At first, when training heterogeneous datasets from different clients, our PMFL algorithm consistently outperforms all three other cases by converging to higher AUC with fewer epochs. Also, even if MetaFL algorithm couldn't perform as well as our PMFL algorithm, its performance is evidently better than w/o FL and FL. Interestingly, sometimes after the FL pretraining process, the performance would be even worse than w/o FL. It indicates the heterogeneity of different clients may hurt the training process of federated learning, and the knowledge from one task would possibly be useless or even malignant for another different task. Table \ref{tab:2} shows this result more explicitly. Take Pleural Other for example, training directly (w/o FL) after 10 epochs achieves a final AUC of $0.8702$, whereas FL only obtained the AUC of $0.7166$ which is worse than training directly, MetaFL could obtain a better AUC of $0.9489$, and PMFL could further improve the AUC to be $0.9799$. 
\begin{figure}[ht]
\centering
\includegraphics[width=\columnwidth]{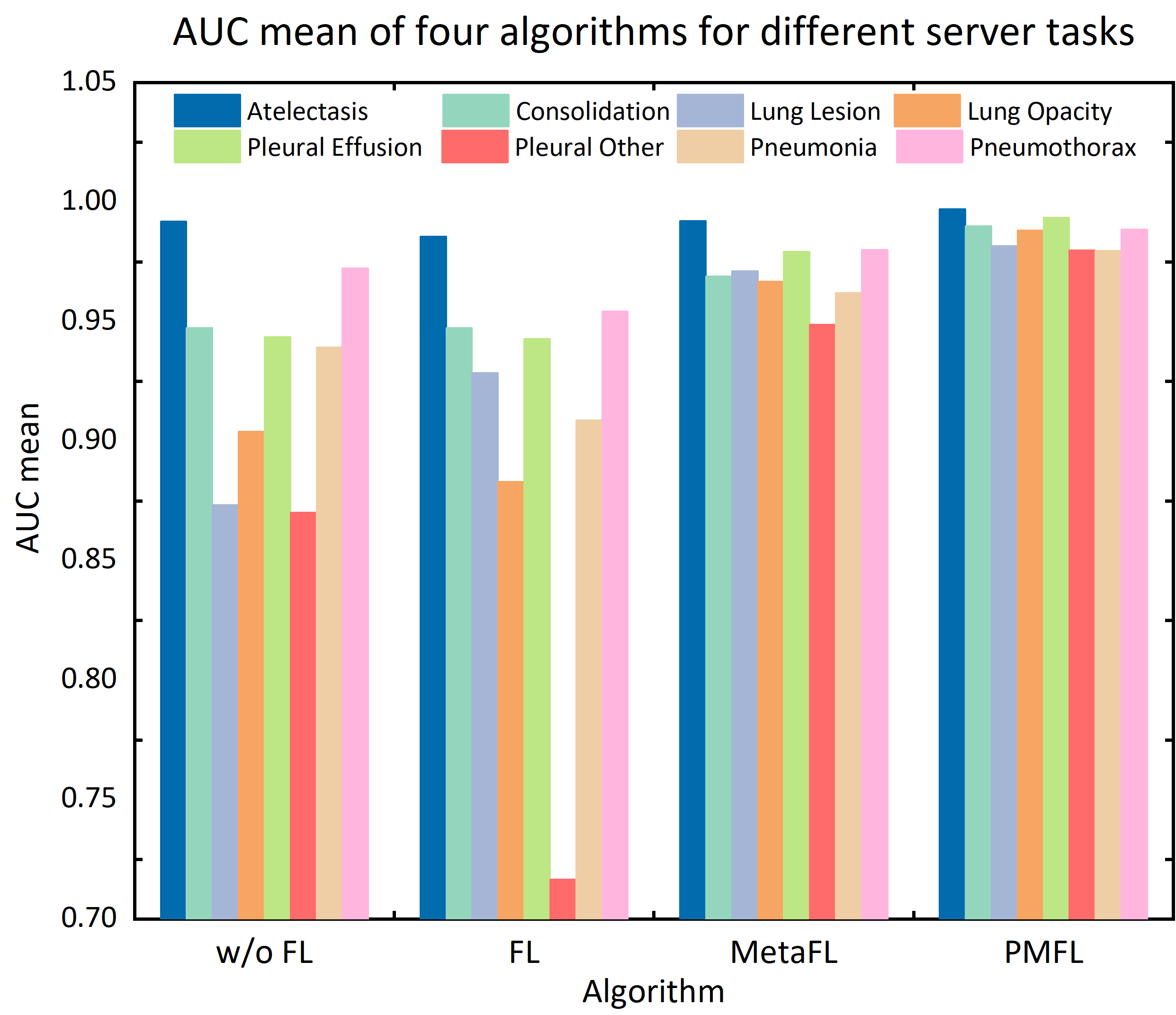}
\caption{Comparison of AUC mean of four algorithms for 8 different server tasks.}
\label{fig:mean}
\end{figure}

Second, PMFL converged not only faster but also more stable. In Fig. \ref{fig:auc}, the thickness of the line represents the standard deviation of the AUC in 5 different experiments. The training process of our PMFL algorithm could converge very fast, mostly in less than 4 epochs. In addition, The red line in the image denotes PMFL algorithm and it's grossly thinner than the other three lines, especially after training a few epochs. In Table \ref{tab:2}, similarly, when we selected Pleural Other as test task, the standard deviation of the AUC after training 10 epochs is $0.0544$ for training directly, while it's $0.0671$ for FL, $0.0554$ for MetaFL, and $0.0036$ for PMFL. 

Third, we could conclude that the improvement of our algorithm is different for various training tasks and test tasks. For instance, when choosing Lung Lesion as test task, training directly could achieve the AUC of $0.8733$, and the MetaFL algorithm could improve it to be $0.9713$ while the PMFL algorithm could further improve it to be $0.9817$; nevertheless, when keeping the training tasks the same but choosing Pleural Other as test task, training directly obtained the AUC of $0.8702$, and MetaFL could improve it to be $0.9489$ while PMFL could further improve it to be $0.9799$. Therefore, the improvement for Lung Lesion task is greater. We believe that this is because the five training tasks are more similar to Lung Lesion when compared with Pleural Other. To show this more persuasively, we also try to treat Lung Lesion as test task but selected five different training tasks and we find that the improvements of MetaFL and PMFL also change.

Besides, we also compare the performance of PMFL algorithm with 3 clients and 5 clients. In Fig. \ref{fig:3v5}, the blue line shows the result of 3-clients while the red line denotes 5-clients case and the black line is the result of training directly. At first, we could conclude that the final performance after 10 epochs of training with 3-clients PMFL is extremely similar to the 5-clients PMFL's. However, the standard deviation of the blue line is smaller. It means that 3-clients PMFL is more stable than 5-clients PMFL. The results from Table \ref{tab:3} show this more clearly. Both PMFL algorithms perform better than training directly, and their final ROC AUC scores are almost the same while the standard deviation of 3-clients PMFL is $0.0029$ which is slightly smaller than $0.0033$ of 5-clients. And this also makes sense, because when you pretrain the model with more different tasks, the heterogeneous property would make a bigger influence.
\begin{figure}[!ht]
\centering
\includegraphics[width=\columnwidth]{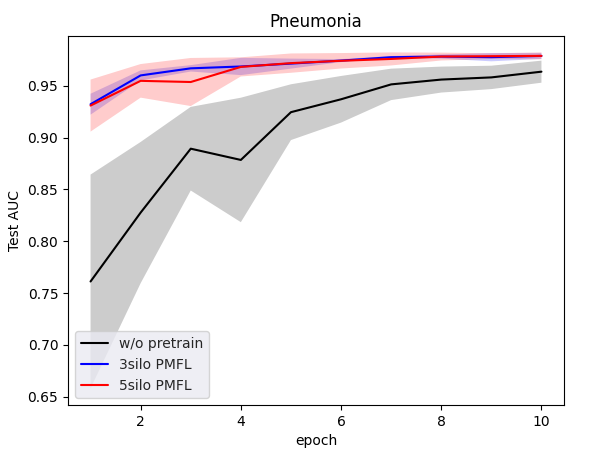}
\caption{Plot of average AUC score for without pretrain, with 3-clients PMFL, with 5-clients PMFL when choosing Pneumonia as test task. The lines in this figure show the average AUC of 5 experiments while the thickness of the lines shows standard deviation for each case.}
\label{fig:3v5}
\end{figure}

\begin{table}[!ht]
\caption{Compare ROC AUC for training directly without any pretrain process, PMFL with 3 clients, and PMFL with 5 clients when treating Pneumonia as target test task.}
\centering
\begin{tabular}{l l l l}
\hline
Test task & Algorithm & ROC AUC \\
\hline
\multirow{4}{6em}{Pneumonia}
& w/o pretrain & $0.9636\pm0.0106$\\

& 3-clients PMFL & $0.9788\pm0.0029$\\

& 5-clients PMFL & $0.9789\pm0.0033$\\

\hline
\end{tabular}

\label{tab:3}
\end{table}

\subsection{eICU}
\textbf{Dataset.}
The eICU collaborative Reseach Database, collected through the Philips eICU program, contains highly granular critical care data of 200,859 patients admitted to 208 hospitals from across the United States \cite{pollard2018eicu}. For the purpose of our study, we mainly utilized three files: admissionDrug.csv.gz, patient.csv.gz and admissionDx.csv.gz.


With these three files, we could obtain the disease labels for each patient from the \textit{apacheadmissiondx} variable in the \textit{patient.csv.gz} file. In order to generate the diagnosis text data for each patient, we extract all the drug names for this patient from the \textit{admissionDrug.csv.gz} file and the diagnosis data from the \textit{admissionDx.csv.gz} table, and integrated them into a single text file for this patient. Finally, we got 40678 effective patient samples for our research, and 7 different disease labels for each sample. In order to generate the heterogeneous tasks for our experiment, with the same scheme of MIMIC-CXR v2.0.0, we extract 5 silo datasets from the original dataset. After that, we would select three different clients for training tasks to pretrain the traditional federated learning, MetaFL, and PMFL models, and select another one for test task to evaluate in the server.

\textbf{Model and performance.}
Similar to the model for MIMIC dataset, a simple LSTM model worked as our meta model. To show better performance for this dataset, we decrease the batch size to half of the MIMIC dataset while all the other hyperparameters are the same. 

\begin{table*}[!hbt]
\caption{Compare the results of eICU for four different algorithms with the same metrics. In this table, Precision, Recall and $ F_1$ score are obtained with the threshold which maximizes Youden's Index.}
\centering
\begin{tabular}{l l l l l l}
\hline
Dataset & Algorithm & ROC AUC & Precision & Recall & $ F_1$ score \\
\hline
\multirow{4}{6em}{eICU} 
& w/o FL & $0.9218\pm0.0217$ & $0.7296\pm0.0821$ & $0.8733\pm0.0814$ & $0.7877\pm0.0294$\\

& FL & $0.8774\pm0.0524$ & $0.7160\pm0.0662$ & $0.7867\pm0.1035$ & $0.7461\pm0.0672$\\

& MetaFL & $0.9299\pm0.0181$ & $0.7226\pm0.0556$ & $0.9033\pm0.0823$ & $0.7983\pm0.0328$\\

& \textbf{PMFL(Our method)} & $\mathbf{0.9562\pmb{\pm}0.0090}$ & $\mathbf{0.7671\pmb{\pm}0.0855}$ & $\mathbf{0.9333\pmb{\pm}0.0667}$ & $\mathbf{0.8358\pmb{\pm}0.0235}$\\
\hline
\end{tabular}
\label{tab:5}
\end{table*}

For this dataset, we also compare our PMFL algorithm with w/o FL, FL and the MetaFL algorithm with the same metrics. In Table \ref{tab:5}, this is the result of eICU dataset when treating CHF, pneumonia, and pulmonary as the training task labels for clients and treating CVA as the test task label for the server evaluation. At first, we could find that compared with w/o FL, regular federated learning (FL) couldn't work for heterogeneous datasets but worsened the training performance for our eICU dataset. Actually, it performs largely worse for this dataset when compared with MIMIC-CXR v2.0.0 dataset, and this is because the tasks for all clients are more different. In addition, our algorithm still outperforms all the other three methods, and we could observe that our PMFL algorithm could improve the AUC of training from 0.9218 to 0.9562 after training 10 epochs, which is better than FL and MetaFL algorithms. Also, the standard deviation of our PMFL shows a more competitive result while MetaFL slightly improves the performance of w/o FL. However, for FL, the result is significantly terrible. Not only does the AUC score decrease to 0.8774, the standard deviation also increases from 0.0217 to 0.0524. Also, for the precision, recall, and $ F_1$ score at the threshold when maximizing the Youden's index, our PMFL algorithm still shows the best result.

By comparing the results of MIMIC-CXR v2.0.0 and eICU datasets, we could find that the performance of regular federated learning would be influenced a lot by the heterogeneity while our proposed algorithm could handle it well. With more different training tasks, regular federated learning would perform worse.


\section{Conclusion}
By effectively exploiting the large amount of data coming from mobile devices, more robust machine learning models would be achieved. While traditional Federated learning, like FedAvg algorithm, has been proven that it's really helpful for homogeneous training with distributed data, it's not suitable for heterogeneous training tasks. In order to successfully apply federated learning to heterogeneous scenarios, we propose a new federated learning algorithm, which would feedback the training losses rather than model parameters to the server after the local update, inspired by meta-learning. And, we have confirmed that our new federated learning algorithm (MetaFL) significantly outperforms conventional federated learning for heterogeneous tasks. It achieves not only better ROC AUC or other scores but the faster training speed. Furthermore, due to the advantage of transfer learning for model generalization, the integration of transfer learning and our algorithm, which is partial meta-federated learning (PMFL), further improves the model performance for the two medical datasets when compared with MetaFL.

In this paper, we find that the performance of the normal federated learning would decrease more while our proposed MetaFL and PMFL algorithms perform more stably, when the divergence of tasks from different clients gets bigger. Nevertheless, we believe that there is some limit for our proposed algorithms. And if the divergence between different tasks goes beyond this limit, our algorithms would also perform dreadfully. Therefore, future research could focus on exploring this limit for our algorithms. To be specific, we could define the distance between different tasks. The bigger distance between the two tasks means that these two tasks are more different. How to compute this distance is also a promising future research direction. After we figure out how to compute this distance, we could explore the relationship between this distance and the performance of all different algorithms. Also, the limit for our proposed MetaFL and PMFL algorithms could be obtained. In addition, if the distance between tasks is extremely small, we believe that the regular federated learning algorithm would also perform very well. Based on this, for different clients, we could quickly judge whether their datasets could be utilized by regular federated learning. Further, we could know the most suitable algorithm for different cases with this distance.

Besides, during our experiments, we try to generate clients with similar sizes. But what if the size is hugely different? Thus, the other viable research direction is to explore the influence of imbalanced datasets. In this case, maybe the training task of the client which includes the most samples would dominate other clients whose datasets are considerably small.

\section*{Acknowledgments}
We are sincerely grateful to the MIT Laboratory for Computational Physiology for providing the MIMIC-CXR dataset. We are also grateful to Philips Health Care and MIT Lab for Computational Physiology for the provision of the eICU dataset. The financial supports of the “the Fundamental Research Funds for the Central Universities” (2021JBM001) is acknowledged.





\bibliographystyle{IEEEtran}
\bibliography{IEEEabrv, main}

\begin{thebibliography}{10}
\providecommand{\url}[1]{#1}
\csname url@samestyle\endcsname
\providecommand{\newblock}{\relax}
\providecommand{\bibinfo}[2]{#2}
\providecommand{\BIBentrySTDinterwordspacing}{\spaceskip=0pt\relax}
\providecommand{\BIBentryALTinterwordstretchfactor}{4}
\providecommand{\BIBentryALTinterwordspacing}{\spaceskip=\fontdimen2\font plus
\BIBentryALTinterwordstretchfactor\fontdimen3\font minus \fontdimen4\font\relax}
\providecommand{\BIBforeignlanguage}[2]{{%
\expandafter\ifx\csname l@#1\endcsname\relax
\typeout{** WARNING: IEEEtran.bst: No hyphenation pattern has been}%
\typeout{** loaded for the language `#1'. Using the pattern for}%
\typeout{** the default language instead.}%
\else
\language=\csname l@#1\endcsname
\fi
#2}}
\providecommand{\BIBdecl}{\relax}
\BIBdecl

\bibitem{McMahan2017CommunicationEfficient}
H.~B. McMahan, E.~Moore, D.~Ramage, S.~Hampson, and B.~A. y~Arcas, ``Communication-efficient learning of deep networks from decentralized data,'' 2017.

\bibitem{JK2016federated}
J.~Konečný, H.~B. McMahan, D.~Ramage, and P.~Richtárik, ``Federated optimization: Distributed machine learning for on-device intelligence,'' 2016.

\bibitem{Bonawitz2017}
\BIBentryALTinterwordspacing
K.~Bonawitz, V.~Ivanov, B.~Kreuter, A.~Marcedone, H.~B. McMahan, S.~Patel, D.~Ramage, A.~Segal, and K.~Seth, ``Practical secure aggregation for privacy-preserving machine learning,'' in \emph{Proceedings of the 2017 {ACM} {SIGSAC} Conference on Computer and Communications Security}.\hskip 1em plus 0.5em minus 0.4em\relax {ACM}, Oct. 2017. [Online]. Available: \url{https://doi.org/10.1145/3133956.3133982}
\BIBentrySTDinterwordspacing

\bibitem{JK2017federated}
J.~Konečný, H.~B. McMahan, F.~X. Yu, P.~Richtárik, A.~T. Suresh, and D.~Bacon, ``Federated learning: Strategies for improving communication efficiency,'' 2017.

\bibitem{chen2019federated}
F.~Chen, M.~Luo, Z.~Dong, Z.~Li, and X.~He, ``Federated meta-learning with fast convergence and efficient communication,'' 2019.

\bibitem{10.5555/3294996.3295196}
V.~Smith, C.-K. Chiang, M.~Sanjabi, and A.~Talwalkar, ``Federated multi-task learning,'' in \emph{Proceedings of the 31st International Conference on Neural Information Processing Systems}, ser. NIPS'17.\hskip 1em plus 0.5em minus 0.4em\relax Red Hook, NY, USA: Curran Associates Inc., 2017, p. 4427–4437.

\bibitem{zhao2018federated}
Y.~Zhao, M.~Li, L.~Lai, N.~Suda, D.~Civin, and V.~Chandra, ``Federated learning with non-iid data,'' 2018.

\bibitem{geyer2018differentially}
R.~C. Geyer, T.~Klein, and M.~Nabi, ``Differentially private federated learning: A client level perspective,'' 2018.

\bibitem{10.1145/3298981}
\BIBentryALTinterwordspacing
Q.~Yang, Y.~Liu, T.~Chen, and Y.~Tong, ``Federated machine learning: Concept and applications,'' vol.~10, no.~2, 2019. [Online]. Available: \url{https://doi.org/10.1145/3298981}
\BIBentrySTDinterwordspacing

\bibitem{lin2020deep}
Y.~Lin, S.~Han, H.~Mao, Y.~Wang, and W.~J. Dally, ``Deep gradient compression: Reducing the communication bandwidth for distributed training,'' 2020.

\bibitem{schmidhuber:1987:srl}
\BIBentryALTinterwordspacing
J.~Schmidhuber, ``Evolutionary principles in self-referential learning. on learning now to learn: The meta-meta-meta...-hook,'' Diploma Thesis, Technische Universitat Munchen, Germany, 1987. [Online]. Available: \url{http://www.idsia.ch/~juergen/diploma.html}
\BIBentrySTDinterwordspacing

\bibitem{Bengio97onthe}
S.~Bengio, Y.~Bengio, J.~Cloutier, and J.~Gecsei, ``On the optimization of a synaptic learning rule,'' 1997.

\bibitem{finn2018metalearning}
C.~Finn and S.~Levine, ``Meta-learning and universality: Deep representations and gradient descent can approximate any learning algorithm,'' 2018.

\bibitem{finn2017modelagnostic}
C.~Finn, P.~Abbeel, and S.~Levine, ``Model-agnostic meta-learning for fast adaptation of deep networks,'' 2017.

\bibitem{Rajeswaran2019MetaLearningWI}
A.~Rajeswaran, C.~Finn, S.~Kakade, and S.~Levine, ``Meta-learning with implicit gradients,'' in \emph{NeurIPS}, 2019.

\bibitem{hospedales2020metalearning}
T.~Hospedales, A.~Antoniou, P.~Micaelli, and A.~Storkey, ``Meta-learning in neural networks: A survey,'' 2020.

\bibitem{Vanschoren2018}
J.~Vanschoren, ``Meta-learning: A survey,'' 10 2018.

\bibitem{10.5555/3157382.3157504}
O.~Vinyals, C.~Blundell, T.~Lillicrap, K.~Kavukcuoglu, and D.~Wierstra, ``Matching networks for one shot learning,'' in \emph{Proceedings of the 30th International Conference on Neural Information Processing Systems}, ser. NIPS'16.\hskip 1em plus 0.5em minus 0.4em\relax Curran Associates Inc., 2016, p. 3637–3645.

\bibitem{10.5555/3045390.3045585}
A.~Santoro, S.~Bartunov, M.~Botvinick, D.~Wierstra, and T.~Lillicrap, ``Meta-learning with memory-augmented neural networks,'' in \emph{Proceedings of the 33rd International Conference on International Conference on Machine Learning - Volume 48}, ser. ICML'16.\hskip 1em plus 0.5em minus 0.4em\relax JMLR.org, 2016, p. 1842–1850.

\bibitem{graves2014neural}
A.~Graves, G.~Wayne, and I.~Danihelka, ``Neural turing machines,'' 2014.

\bibitem{Ravi2017OptimizationAA}
S.~Ravi and H.~Larochelle, ``Optimization as a model for few-shot learning,'' in \emph{ICLR}, 2017.

\bibitem{pmlr-v70-munkhdalai17a}
\BIBentryALTinterwordspacing
T.~Munkhdalai and H.~Yu, ``Meta networks,'' in \emph{Proceedings of the 34th International Conference on Machine Learning}, ser. Proceedings of Machine Learning Research, D.~Precup and Y.~W. Teh, Eds., vol.~70.\hskip 1em plus 0.5em minus 0.4em\relax PMLR, 06--11 Aug 2017, pp. 2554--2563. [Online]. Available: \url{http://proceedings.mlr.press/v70/munkhdalai17a.html}
\BIBentrySTDinterwordspacing

\bibitem{Sadilek2020.12.22.20245407}
\BIBentryALTinterwordspacing
A.~Sadilek, L.~Liu, D.~Nguyen, M.~Kamruzzaman, B.~Rader, A.~Ingerman, S.~Mellem, P.~Kairouz, E.~O. Nsoesie, J.~MacFarlane, A.~Vullikanti, M.~Marathe, P.~Eastham, J.~S. Brownstein, M.~Howell, and J.~Hernandez, ``Privacy-first health research with federated learning,'' \emph{medRxiv}, 2020. [Online]. Available: \url{https://www.medrxiv.org/content/early/2020/12/24/2020.12.22.20245407}
\BIBentrySTDinterwordspacing

\bibitem{info:doi/10.2196/20891}
\BIBentryALTinterwordspacing
G.~H. Lee and S.-Y. Shin, ``Federated learning on clinical benchmark data: Performance assessment,'' \emph{J Med Internet Res}, vol.~22, no.~10, p. e20891, Oct 2020. [Online]. Available: \url{http://www.jmir.org/2020/10/e20891/}
\BIBentrySTDinterwordspacing

\bibitem{boughorbel2019federated}
S.~Boughorbel, F.~Jarray, N.~Venugopal, S.~Moosa, H.~Elhadi, and M.~Makhlouf, ``Federated uncertainty-aware learning for distributed hospital ehr data,'' 2019.

\bibitem{xu2020federated}
J.~Xu, B.~S. Glicksberg, C.~Su, P.~Walker, J.~Bian, and F.~Wang, ``Federated learning for healthcare informatics,'' 2020.

\bibitem{liu2022confederated}
D.~Liu, K.~Fox, G.~Weber, and T.~Miller, ``Confederated learning in healthcare: training machine learning models using disconnected data separated by individual, data type and identity for large-scale health system intelligence,'' \emph{Journal of Biomedical Informatics}, p. 104151, 2022.

\bibitem{10.1145/3412357}
\BIBentryALTinterwordspacing
B.~Pfitzner, N.~Steckhan, and B.~Arnrich, ``Federated learning in a medical context: A systematic literature review,'' \emph{ACM Trans. Internet Technol.}, vol.~21, no.~2, Jun. 2021. [Online]. Available: \url{https://doi.org/10.1145/3412357}
\BIBentrySTDinterwordspacing

\bibitem{osti_10064144}
\BIBentryALTinterwordspacing
J.~Lee, J.~Sun, F.~Wang, S.~Wang, C.-H. Jun, and X.~Jiang, ``Privacy-preserving patient similarity learning in a federated environment: Development and analysis,'' \emph{JMIR Medical Informatics}, vol.~6, no.~2, 2018. [Online]. Available: \url{https://par.nsf.gov/biblio/10064144}
\BIBentrySTDinterwordspacing

\bibitem{huang2019patient}
L.~Huang and D.~Liu, ``Patient clustering improves efficiency of federated machine learning to predict mortality and hospital stay time using distributed electronic medical records,'' 2019.

\bibitem{liu2019twostage}
D.~Liu, D.~Dligach, and T.~Miller, ``Two-stage federated phenotyping and patient representation learning,'' 2019.

\bibitem{liu2018fadlfederatedautonomous}
D.~Liu, T.~Miller, R.~Sayeed, and K.~D. Mandl, ``Fadl:federated-autonomous deep learning for distributed electronic health record,'' 2018.

\bibitem{BRISIMI201859}
\BIBentryALTinterwordspacing
T.~S. Brisimi, R.~Chen, T.~Mela, A.~Olshevsky, I.~C. Paschalidis, and W.~Shi, ``Federated learning of predictive models from federated electronic health records,'' \emph{International Journal of Medical Informatics}, vol. 112, pp. 59--67, 2018. [Online]. Available: \url{https://www.sciencedirect.com/science/article/pii/S138650561830008X}
\BIBentrySTDinterwordspacing

\bibitem{jimaging7020031}
\BIBentryALTinterwordspacing
P.~Zhang, J.~Li, Y.~Wang, and J.~Pan, ``Domain adaptation for medical image segmentation: A meta-learning method,'' \emph{Journal of Imaging}, vol.~7, no.~2, 2021. [Online]. Available: \url{https://www.mdpi.com/2313-433X/7/2/31}
\BIBentrySTDinterwordspacing

\bibitem{park2021metalearning}
H.~Park, G.~M. Lee, S.~Kim, G.~H. Ryu, A.~Jeong, S.~H. Park, and M.~Sagong, ``A meta-learning approach for medical image registration,'' 2021.

\bibitem{Hu2018MetaLearningFM}
S.~Hu, J.~M. Tomczak, and M.~Welling, ``Meta-learning for medical image classification,'' in \emph{1st Conference on Medical Imaging with Deep Learning (MIDL 2018)}, 2018.

\bibitem{9150592}
K.~Mahajan, M.~Sharma, and L.~Vig, ``Meta-dermdiagnosis: Few-shot skin disease identification using meta-learning,'' in \emph{2020 IEEE/CVF Conference on Computer Vision and Pattern Recognition Workshops (CVPRW)}, 2020, pp. 3142--3151.

\bibitem{nichol2018firstorder}
A.~Nichol, J.~Achiam, and J.~Schulman, ``On first-order meta-learning algorithms,'' 2018.

\bibitem{khadga2021fewshot}
R.~Khadga, D.~Jha, S.~Ali, S.~Hicks, V.~Thambawita, M.~A. Riegler, and P.~Halvorsen, ``Few-shot segmentation of medical images based on meta-learning with implicit gradients,'' 2021.

\bibitem{chen2021metadelta}
Y.~Chen, C.~Guan, Z.~Wei, X.~Wang, and W.~Zhu, ``Metadelta: A meta-learning system for few-shot image classification,'' 2021.

\bibitem{fallah2020personalized}
A.~Fallah, A.~Mokhtari, and A.~Ozdaglar, ``Personalized federated learning: A meta-learning approach,'' \emph{arXiv preprint arXiv:2002.07948}, 2020.

\bibitem{yoon2017lifelong}
J.~Yoon, E.~Yang, J.~Lee, and S.~J. Hwang, ``Lifelong learning with dynamically expandable networks,'' \emph{arXiv preprint arXiv:1708.01547}, 2017.

\bibitem{golkar2019continual}
S.~Golkar, M.~Kagan, and K.~Cho, ``Continual learning via neural pruning,'' \emph{arXiv preprint arXiv:1903.04476}, 2019.

\bibitem{johnson2019mimic}
A.~Johnson, T.~Pollard, R.~Mark, S.~Berkowitz, and S.~Horng, ``Mimic-cxr database,'' \emph{PhysioNet https://doi. org/10.13026/C2JT1Q}, 2019.

\bibitem{johnson2019mimicpaper}
A.~E. Johnson, T.~J. Pollard, N.~R. Greenbaum, M.~P. Lungren, C.-y. Deng, Y.~Peng, Z.~Lu, R.~G. Mark, S.~J. Berkowitz, and S.~Horng, ``Mimic-cxr-jpg, a large publicly available database of labeled chest radiographs,'' \emph{arXiv preprint arXiv:1901.07042}, 2019.

\bibitem{goldberger2000physiobank}
A.~L. Goldberger, L.~A. Amaral, L.~Glass, J.~M. Hausdorff, P.~C. Ivanov, R.~G. Mark, J.~E. Mietus, G.~B. Moody, C.-K. Peng, and H.~E. Stanley, ``Physiobank, physiotoolkit, and physionet: components of a new research resource for complex physiologic signals,'' \emph{circulation}, vol. 101, no.~23, pp. e215--e220, 2000.

\bibitem{johnsonmimic}
A.~Johnson, M.~Lungren, Y.~Peng, Z.~Lu, R.~Mark, S.~Berkowitz, and S.~Horng, ``Mimic-cxr-jpg-chest radiographs with structured labels.''

\bibitem{pollard2018eicu}
T.~J. Pollard, A.~E. Johnson, J.~D. Raffa, L.~A. Celi, R.~G. Mark, and O.~Badawi, ``The eicu collaborative research database, a freely available multi-center database for critical care research,'' \emph{Scientific data}, vol.~5, no.~1, pp. 1--13, 2018.

\end{thebibliography}
 

\end{document}